%% file: root.tex
\documentclass[letterpaper, 10pt, conference]{ieeeconf}
\IEEEoverridecommandlockouts 
\usepackage{subcaption}
\usepackage[font=small]{caption}
\usepackage{subcaption}
\usepackage{tikz}
\usepackage{xcolor}
\usepackage{graphicx}
\usepackage{amsmath,amssymb}
\usepackage[ruled,vlined,linesnumbered]{algorithm2e}
\SetKwFor{For}{for}{do}{end for}
\usepackage{stfloats} 
\usepackage[utf8]{inputenc}

\let\labelindent\relax
 
\usepackage{amsthm}
\usepackage{amsfonts}
\usepackage{amsmath}
\usepackage{amssymb}

\usepackage{float}

\newtheorem{proposition}{Proposition}
\usepackage{amsthm}
\newtheorem{remark}{Remark}
\makeatletter
\let\NAT@parse\undefined
\makeatother
\usepackage{url}
\usepackage[colorlinks,citecolor=blue,linkcolor=red]{hyperref}
\usepackage{cite}
\usepackage{cleveref}
\crefrangeformat{equation}{(#3#1#4)--(#5#2#6)}
\usepackage{import}
\usepackage{enumitem}

\usepackage{booktabs}
\title{\LARGE \bf PAD-TRO: Projection-Augmented Diffusion for Direct Trajectory Optimization}

\author{
    Jushan Chen and 
    Santiago Paternain
    \thanks{The authors are with the Department of Electrical, Computer, and Systems Engineering, Rensselaer Polytechnic Institute, 110 8th St, Troy, NY 12180, USA. {\tt\small \{chenj72, paters\}@rpi.edu}}
}

\begin{document}

\setlength{\textfloatsep}{5pt plus 2pt minus 4pt}
\setlength{\dbltextfloatsep}{3pt}
\setlist{nosep} 

\newcommand \blue[1]         {{\color{blue}#1}}
\newcommand\green[1]       {{\color[rgb]{0.10,0.50,0.10}#1}}
\newcommand \red[1]         {{\color{red}#1}}

\maketitle
\thispagestyle{empty}
\pagestyle{empty}

\import{./}{Abstract/abstract.tex}

\section{INTRODUCTION} \label{sec:introduction}
\import{./}{Introduction/introduction.tex}

\section{PROBLEM FORMULATION} \label{sec:problem_formulation}
\import{./}{ProblemFormulation/problem.tex}

\section{Background on diffusion for Trajectory Optimization} \label{sec:background}
\import{./}{Background/background.tex}

\section{CONSTRAINED DIFFUSION FOR DIRECT TRAJECTORY OPTIMIZATION}
\label{sec:methodology}
\import{./}{Methodology/methodology.tex}

\section{Simulation Studies}
\label{sec:results}
\import{./}{Experiments/results.tex}

\section{Conclusion}\label{sec:conclusion}
We have proposed a novel direct trajectory optimization framework via model-based diffusion. We have introduced a novel gradient-free projection mechanism for ensuring dynamic feasibility. To balance exploration and optimality, we have adopted a novel bi-level noise schedule on both the diffusion horizon and the trajectory prediction horizon. Our results show that our method outperforms baselines in various metrics in a quadrotor navigation problem in a cluttered environment. The projection framework in our approach relies on pre-set noise thresholds. Further investigation is required to study how to design an adaptive and potentially more efficient projection mechanism that is dynamics-aware. In addition, accelerating the projection-augmented diffusion process for faster convergence is a potential research direction. Last but not least, further validation in hardware experiments on complex robotic systems, such as quadruped robots, is needed.

\bibliographystyle{ieeetr}
\bibliography{references}

\end{document}

%% file: Abstract/abstract.tex
\begin{abstract}
Recently, diffusion models have gained popularity and attention in trajectory optimization due to their capability of modeling multi-modal probability distributions. However, addressing nonlinear equality constraints, i.e, dynamic feasibility, remains a great challenge in diffusion-based trajectory optimization. Recent diffusion-based trajectory optimization frameworks rely on a single-shooting style approach where the denoised control sequence is applied to forward propagate the dynamical system, which cannot explicitly enforce constraints on the states and frequently leads to sub-optimal solutions. In this work, we propose a novel direct trajectory optimization approach via model-based diffusion, which directly generates a sequence of states. To ensure dynamic feasibility, we propose a gradient-free projection mechanism that is incorporated into the reverse diffusion process. Our results show that, compared to a recent state-of-the-art baseline, our approach leads to zero dynamic feasibility error and approximately 4x higher success rate in a quadrotor waypoint navigation scenario involving dense static obstacles. 
\end{abstract}

%% file: Introduction/introduction.tex
Trajectory optimization is a critical task in robotics. Traditional methods in trajectory optimization transcribe the original continuous-time optimization problem into a corresponding nonlinear programming problem (NLP) via shooting methods or collocation methods. In particular, shooting-based methods use the control variables as optimization variables and enforce dynamic feasibility via forward propagation, whereas collocation methods are entirely based on function approximation and use both control and state variables as the optimization variables \cite{kelly_trajectory_opt_tutorial}. Solving such NLPs requires gradient-based optimization \cite{sequential_convex_fleet,quadrotor_generation_gao_Fei} such as interior-point solvers \cite{Byrd1999AnIP}, which are often not robust against nonconvex objectives or constraints and might be stuck in local minima. In contrast, sampling based trajectory optimization methods are, in general, more robust to local minima, such as the Model Predictive Path Integral \cite{MPPI} framework and Cross Entropy motion planning \cite{CEM}, but the downside is that the quality of solutions is highly dependent on the scale of Gaussian noise, which might lead to over-approximation or under-approximation. On the other hand, geometric path planning algorithms such as RRT \cite{LaValle1998RapidlyexploringRT} and the many variants that follow from RRT, are efficient at finding a global path, but they disregard the problem-specific constraints and dynamic feasibility for robotic systems.

Recently, diffusion models have been widely adopted in computer vision tasks such as image and video generation \cite{Ho2020DenoisingDP,DDIM}. Due to its expressive power of modeling multi-modal distributions, several recent works have incorporated diffusion in trajectory optimization or control policy learning in robotics \cite{Bouvier2025DDATDP,kurtz2024equalityconstraineddiffusion,dialMPC,zhang2025constraineddiffusers,diffusion_policy,safe_diffuser}. In particular, these diffusion-style methods can be divided into two categories: learning-based and model-based. Learning-based trajectory planning frameworks \cite{Bouvier2025DDATDP,zhang2025constraineddiffusers,diffusion_policy,safe_diffuser} rely on expert demonstration data to train a noise prediction network. Then, trajectory generation is performed in a reverse diffusion process. They resemble the classic Denoising Diffusion Probabilistic Models (DDPM) \cite{Ho2020DenoisingDP}. In contrast, model-based diffusion frameworks \cite{pan2024modelbaseddiffusion,kurtz2024equalityconstraineddiffusion, dialMPC} assume that we have some knowledge about the target distribution of optimal trajectories, and rely on Langevin sampling while estimating the score function, i.e., the gradient of the log-likelihood. For example, \cite{pan2024modelbaseddiffusion,kurtz2024equalityconstraineddiffusion,dialMPC} assume that the target distribution of the optimal trajectories follows a Boltzmann distribution. Intuitively, the cost function is optimized by sampling from the corresponding Boltzmann distribution by annealing the temperature coefficient to zero \cite{pan2024modelbaseddiffusion}. Model-Based Diffusion (MBD) \cite{pan2024modelbaseddiffusion} solves the trajectory optimization problem by generating a sequence of control actions at each step of the reverse diffusion process. Then, the sequence of control actions in the last reverse diffusion step is applied to forward propagate the system dynamics, leading to a sequence of dynamically feasible states. In other words, MBD \cite{pan2024modelbaseddiffusion} is analogous to single shooting \cite{altro} in classic trajectory optimization. In contrast, Equality Constrained Diffusion for Direct Trajectory Optimization (DRAX) \cite{kurtz2024equalityconstraineddiffusion} generates both a sequence of states and actions via diffusion, and enforces soft dynamic feasibility by using an augmented Lagrangian term to penalize violation of dynamic feasibility. To the best of the authors' knowledge, \cite{kurtz2024equalityconstraineddiffusion} is the only recent work that attempts to address nonlinear equality constraints in a model-based diffusion framework for direct trajectory optimization. 

However, both MBD \cite{pan2024modelbaseddiffusion} and DRAX \cite{kurtz2024equalityconstraineddiffusion} suffer from critical drawbacks: we empirically discover that for a high-dimensional dynamical system (e.g., a quadrotor) navigating through a cluttered environment, MBD \cite{pan2024modelbaseddiffusion} cannot achieve close convergence to the goal position (see Section \ref{sec:results}). On the other hand, DRAX is more robust to local minima. However, due to the soft penalty on the dynamic feasibility violation, the optimized trajectory leads to a high dynamic feasibility violation, potentially inhibiting a low-level controller from trajectory tracking. To address the aforementioned limitations, we propose PAD-TRO, \textit{\underline{P}rojection-\underline{A}ugmented \underline{D}iffusion for Direct \underline{Tr}ajectory \underline{O}ptimization}. The main contributions of our work are summarized as follows: 
\begin{itemize}
    \item[(i)] We propose a novel model-based diffusion algorithm for direct trajectory optimization; 
    \item[(ii)] We integrate a gradient-free projection mechanism to address nonlinear equality constraints (i.e., dynamic feasibility) during the reverse diffusion process;
    \item[(iii)] We show that our method leads to exact convergence to the goal position and results in \textbf{zero dynamic feasibility violation} and \textbf{4x higher success rate} compared to a similar baseline \cite{kurtz2024equalityconstraineddiffusion}.
\end{itemize}
The rest of the work is organized as follows: in Section \ref{sec:problem_formulation}, we introduce the trajectory optimization problem, which makes explicit the challenges associated with solving it through sampling. Namely, ensuring dynamic feasibility. In Section \ref{sec:background}, we discuss the theoretical background relevant to diffusion and trajectory optimization. Then, we present our novel diffusion-based trajectory optimization framework (see Section \ref{sec:methodology}). Other than concluding remarks in Section \ref{sec:conclusion}, this work closes with numerical examples where we evaluate the proposed method in a quadrotor navigation scenario in a cluttered environment (see Section \ref{sec:results}).

%% file: ProblemFormulation/problem.tex
We denote the state and the input of a system as $x\in \mathbb{R}^n$ and  $u\in \mathbb{R}^m$, respectively. We further denote the discrete time by subscripts $t\in \mathbb{N}_0$, i.e., $x_t$ and $u_t$ refer to the state and input at time $t$, respectively. We also define for notational convenience $x_{1:T} = \left\{x_t\right\}_{t=1,\ldots,T}$ and $u_{0:T-1}=~\left\{u_t\right\}_{t=0,\ldots,T-1}$.  In addition, we assume that $u_L$ and $u_U$ are the lower and upper bounds of the control input, respectively, and similarly, $x_L$ and $x_U$ refer to the lower and upper bounds of the state.

Let $f:\mathbb{R}^n\times \mathbb{R}^m \to \mathbb{R}^n$ describe the nonlinear dynamics of the system as a function of the state $x$ and control input $u$. We assume there are $N_{obs}$ obstacles and denote $g:\mathbb{R}^n \to \mathbb{R}^{N_{obs}}$ as the collision avoidance constraints of the system as a function of $x$. $g$ can represent, for example, a function related to the distance to an obstacle.  We also define a terminal set $\mathcal{X}_T \in \mathbb{R}^n$ that constrains the state variable at time $T$. Further, denote $J:~\mathbb{R}^n\times \mathbb{R}^m\to \mathbb{R}$ as the cost function to be minimized, e.g., a quadratic trajectory tracking cost. With these definitions, a general trajectory optimization problem in discrete time can be written as 
\begin{equation}
\begin{aligned}
  &\min_{\,x_{1:T},\,u_{0:T-1}}\;J(x_{1:T},u_{0:T-1})\\
  \text{s.t.}\;&x_{t+1}=f(x_t,u_t),\quad t=0,\dots,T-1,\\
                     &g(x_t)\leq 0,\quad t=1,\dots,T,\\
                     &u_L\leq u_t \leq u_U, t = 0, \dots, T-1\\
                     &x_L \leq x_t \leq x_U, t= 1, \dots, T, \\
                     &x_T \in \mathcal{X}_T,
\end{aligned}\label{eqn:optimization_formulation}
\end{equation}
where $\leq$ and $\geq$ denote component-wise inequalities. Note that, since $x_0$ is the given initial condition, it is excluded from the decision variables. In sampling-based optimization, we transcribe the trajectory optimization problem in~\eqref{eqn:optimization_formulation} into a probability sampling problem \cite{pan2024modelbaseddiffusion}. We use lowercase letters to denote real-valued variables, and by a slight abuse of notation, we use uppercase letters to denote corresponding random variables or a batch of real-valued samples. We clarify what uppercase letters refer to whenever necessary. 

While trajectory optimization has a longstanding tradition in solving~\eqref{eqn:optimization_formulation} using a nonlinear optimization solver, e.g., Casadi~\cite{casadi}, it does not always results in an optimal or even feasible solution. Thus, to tackle these challenges, we aim to sample a \textit{target trajectory distribution} $p_0$ which is given by \begin{equation}
    p_0(X ) \propto p_d(X)p_J (X)p_g(X),
\label{eqn:optimal_distribution}
\end{equation}
where $p_d(X) \propto \prod_{t=1}^{T}\mathbf{1}(x_t=f(x_{t-1},u_{t-1}))$ denotes the dynamic feasibility condition, $p_J(X)\propto e^{-J(X)/\lambda}$ denotes optimality of the trajectory, and $p_g(X)\propto\prod_{t=1}^{T}\mathbf{1}\left(g_t(x_t,u_t)\leq0\right)$ denotes inequality constraints satisfaction.  The indicator function in $p_d (\cdot)$ and $p_g(\cdot)$ enforce the dynamic feasibility and obstacle avoidance of the trajectory. Indeed, the indicator function evaluates to 1 if the underlying constraints are satisfied; otherwise, it evaluates to 0. The cost of the trajectory $J$ is transformed into a Boltzmann distribution $p_J(\cdot)$ such that when the exponent is small, we get a value close to 1. The challenge of directly sampling from~\eqref{eqn:optimal_distribution} arises from the high-dimensionality and sparsity of samples that strictly follow the target distribution $p_0(X)$. In particular, the dynamic constraints have zero measure. To tackle this challenge, our approach combines model-based diffusion with a gradient-free projection on an approximate reachable set. Before describing the algorithm in detail in Section \ref{sec:methodology}, we discuss the relevant theoretical background of diffusion in the following section.


%% file: Background/background.tex
Diffusion models originate from the theory of Langevin sampling \cite{langevin_sampling}, which aims to sample from a generic probability distribution $p(Z)$ using the gradient of the log-likelihood function $\nabla_z \log p(Z)$, also known as the score function. 
By convention, the standard Langevin Monte Carlo solves an \textit{unconstrained optimization} problem by generating samples from some \textit{target distribution} \cite{safe_diffuser}
\begin{equation}
    p(Z) \propto e^{-J(Z)}
\end{equation}
via the following Langevin stochastic differential equation (SDE)
\begin{equation}
    dZ = -\nabla_{z}J(Z) + \sqrt{2}dW,
\end{equation}
where $W$ denotes a Wiener process random variable, and the score function $\nabla_z \log p(Z)$ is equivalent to $-\nabla_{z}J(Z)$. In this scenario, the Langevin Dynamics are equivalent to stochastic gradient descent. Intuitively, the score function acts as a guidance term such that the diffusion process gradually pushes samples to regions of low cost.
As discussed in \cite{zhang2025constraineddiffusers}, existing literature commonly applies the following discrete-time approximation to simulate continuous-time Langevin Dynamics, leading to Stochastic Gradient Langevin Dynamics (SGLD):
\begin{equation}
    z_{i-1} = z_i + \frac{\beta_i}{2}\nabla_z \log p(z_i) + \sqrt{\beta_i} \epsilon, \epsilon \sim \mathcal{N}(0,I),
    \label{eqn:langevin_sde}
\end{equation}
where $\beta_t$ is a pre-determined variance schedule. However, applying~\eqref{eqn:langevin_sde} directly to solve the constrained trajectory optimization problem by generating samples from the target distribution $p_0(\cdot)$ in~\eqref{eqn:optimal_distribution} is impractical due to the composition of dynamic feasibility $p_d(\cdot)$ and constraints satisfaction $p_g(\cdot)$. To directly compute the score function is challenging unless the objective has specific forms, such as the Boltzmann distribution $p_J(\cdot)$ where $\nabla \log p_J(\cdot)$ only depends
on the gradient of $J(\cdot)$. In particular, the challenge arises in~\eqref{eqn:optimal_distribution} where the gradient of $p_d(\cdot)$ and $p_g(\cdot)$ are not well-defined. Consider the dynamic feasibility component, $\nabla \log p_d(\cdot)$: $p_d(\cdot)$ evaluates to $0$ when the constraint is violated. Then, $ \log p_d(\cdot)$ becomes undefined, and thus we have no access to the gradient $\nabla \log p_d(\cdot)$.
Thus, an estimation of the score function for~\eqref{eqn:optimal_distribution} is needed. To tackle this challenge, the authors of \cite{pan2024modelbaseddiffusion} proposed a model-based diffusion framework for trajectory optimization through the following variance-preserved reverse diffusion process, also known as the ``multi-step score ascent",
\begin{equation}
    u^{i-1} = \frac{1}{\sqrt{\alpha_i}}(u^i + (1-\bar{\alpha}_i)\nabla_{u^i}\log p_i(u^i)),
    \label{eqn:MBD_reverse}
\end{equation}
where $u^i$ := $u^i_{t=0:T-1}$ denotes the control sequence at the reverse diffusion step $i \in [N, N-1,\dots,1]$. The probability $p_i(\cdot)$ is the outcome of iteratively adding Gaussian noise to the target distribution $p_0(\cdot)$. At $i=N$, the control sequence is sampled from a multi-variate normal distribution  $\mathcal{N}(0,I)$, where  $ I$ denotes the $\mathbb{R}^{T m \times Tm}$ identity matrix. In~\eqref{eqn:MBD_reverse}, $\bar{\alpha}_i=\prod_{k=1}^{i}\alpha_k$, where each $\alpha_k$ is defined as $1-\beta_k$, and $\beta_k$ linearly and uniformly increases from an initial $\beta_0 \ll 1$ to a final $\beta_N \ll 1$. The score function $\nabla_u\log p_i(u^i)$ at each reverse diffusion step $i$ is computed via Monte Carlo estimation via parallel dynamics roll-outs: first, at each reverse diffusion step $i$, a batch of $N_s$ samples of control sequences is drawn as
\begin{equation}
    U^{i} \sim \mathcal{N}\left(\frac{u^i}{\sqrt{\bar{\alpha}_{i-1}}}, \left(\frac{1}{\bar{\alpha}_{i-1}}-1\right)I\right),
    \label{eqn:batch_sampling_mbd}
\end{equation}
where the batch $U^i \in \mathbb{R}^{N_s\times T m}$. 
Then, a weighted average sample is computed as
\begin{equation}
    \bar{u}^{i} = \frac{\sum_{u \in U^i} u p_J(u)p_g(u)}{\sum_{u \in U^i} p_J(u)p_g(u)}.
    \label{eqn:weighted_sample_mean_mbd}
\end{equation}
To evaluate $p_J(u)$ $\propto$ $e^{-J(\cdot)/\lambda}$, the control sequence $u$ is applied to the system to obtain a sequence of dynamically feasible states, which is then evaluated by the cost function $J(x_{1:T},u_{0:T-1})$. Thus, the dynamic feasibility $p_d(\cdot)$ is implicitly enforced. For safety constraints $p_g(\cdot)$, if the drawn samples lead to collisions, then they are excluded from the weighted average sample in~\eqref{eqn:weighted_sample_mean_mbd}. 
Next, the score function is approximated as follows using the weighted average sample above:
\begin{equation}
    \nabla_{u^i} \log p_i(u^i) \approx - \frac{u^i - \sqrt{\bar{\alpha}_i} \, \bar{u}^i}{1 - \bar{\alpha}_i}.
    \label{eqn:score_function_mbd}
\end{equation}
At the end of the reverse diffusion process~\eqref{eqn:MBD_reverse}, the final denoised control sequence $u^0$ is executed to forward propagate the system $x_{t+1}=f(x_t,u_t)$ from $x_{0}$ to obtain a dynamically feasible state trajectory $x_{t=0:T}$. Overall, MBD \cite{pan2024modelbaseddiffusion} applies model-based diffusion to sample from~\eqref{eqn:optimal_distribution}, where only the control $u_{0:T-1}$ are the decision variables. This framework is analogous to single shooting in classic gradient-based trajectory optimization.

%% file: Methodology/methodology.tex
One significant drawback of \cite{pan2024modelbaseddiffusion} is that in waypoint navigation scenarios in a cluttered environment, the optimized trajectory cannot reach the target exactly, leading to sub-optimal solutions. MBD \cite{pan2024modelbaseddiffusion} is unable to achieve close convergence since the terminal constraint $x_T\in\mathcal{X}_T$ cannot be enforced due to its single-shooting-like nature.
In addition, random trajectory rollouts in \cite{pan2024modelbaseddiffusion} lead to many collisions in a cluttered environment, and thus the weighted sample average in the score estimation~\eqref{eqn:score_function_mbd} consists of very sparse samples. Consequently, the score function estimated using these random rollouts constitutes a weak ``guidance signal". In contrast, in Section \ref{sec:methodology_2} we propose a constrained diffusion process for direct trajectory optimization by directly sampling the sequence of \textit{predicted states} $\tilde{x}^i_{t=1:T}$. Due to the fact that diffusion models, either data-drive or model-based, cannot automatically produce dynamically feasible states, we draw inspiration from \cite{Bouvier2025DDATDP} to enforce the dynamic feasibility of the predicted sequence of states via a sequential projection scheme. In contrast to the projection scheme in \cite{Bouvier2025DDATDP}, which requires solving a convex optimization problem to project each predicted state $\tilde{x}_{t+1}$ onto the reachable set of $\tilde{x}_t$, we adopt a sampling-based approach to keep the projection scheme gradient-free. We introduce our diffusion-based trajectory optimization framework in Section \ref{sec:methodology_1} by considering optimality $p_J(\cdot)$ and safety $p_g(\cdot)$. Then, we discuss the details of our projection scheme in Section. \ref{sec:methodology_2} to augment our approach and satisfy dynamic feasibility $p_d(\cdot)$.

\subsection{Diffusion for direct sampling of state trajectory}
\label{sec:methodology_1}
To gradually generate samples that follow the target distribution $p_0(\cdot)$~\eqref{eqn:optimal_distribution}, we propose the following approach built upon \cite{pan2024modelbaseddiffusion} by introducing a novel gradient-free projection scheme and a bi-level noise schedule. In \cite{pan2024modelbaseddiffusion}, the noise schedule, i.e., the standard deviation of the Gaussian~\eqref{eqn:batch_sampling}, $\sigma_{i} := \sqrt{\frac{1-\bar{\alpha}_{i-1}}{\bar{\alpha}_{i-1}}}$ decays only along the diffusion horizon. Inspired by \cite{dialMPC}, we propose a bi-level noise schedule $\sigma_{i,t}$, where the noise not only varies along the diffusion horizon $i=N, N-1, \dots, 1$, but also varies along the trajectory prediction horizon, $t=1,\dots,T$. In practice, we find that this allows for a better trade-off between exploration and reaching optimality in the reverse diffusion process. 

We define the novel noise schedule as follows:
\begin{equation}
    \sigma_{i,t} = \sqrt{\frac{1-\bar{\alpha}_{i-1}}{\bar{\alpha}_{i-1}}} \cdot \delta ^t, i \in [N, \dots, 1], t \in [1,\dots, T],
    \label{eqn:noise_Schedule_ours}
\end{equation}
where $\delta \in (0,1)$ controls the noise along the trajectory prediction horizon. Intuitively, the states at time steps later along the trajectory prediction horizon $t=1,\dots T$ receive lower noise, allowing for effective projection of predicted states onto the reachable sets of earlier states, which we discuss in detail in Section \ref{sec:methodology_2}. In addition, $\bar{\alpha_{i}}$ are defined as in Section \ref{sec:background}. 

While MBD \cite{pan2024modelbaseddiffusion} applies forward rollouts via the control inputs to discard samples that result in collisions per the constraints $p_g(\cdot)$, we apply the following exponential collision avoidance cost:
\begin{equation}
g(x)
= \sum_{i=1}^{N_{\mathrm{obs}}} 
\exp\!\left(-\kappa \,\big(\|\,o(x) - o_{obs,i}\,\|_2^2 - r_i^2\big)\right), \kappa>0,
\label{eqn:collision_avoidance}
\end{equation}
where we denote $o(x)$ as the position of the trajectory and $o_{obs,i}$ is the position of the $i^{th}$ obstacle. We also denote $r_i$ as the radius of the $i^{th}$ obstacle. We drop the time indices here for simplicity. When $p_g(\cdot)$ is combined with optimality $p_J(\cdot)$, the estimated score function is weighted by both functions.  In the reverse diffusion process, this has the effect of pushing the denoised trajectories toward a probability distribution that satisfies both inequality constraints, i.e., staying away from obstacles, and lowering the trajectory cost simultaneously. Furthermore, rather than diffusing the inputs to the system as in \eqref{eqn:MBD_reverse}, we diffuse the states. This allows us to enforce the terminal constraint and to reduce the number of trajectories that collide with the obstacles.

To be formal, we initialize the state trajectory $\tilde{x}^N \in \mathbb{R}^{T\times n} \sim \mathcal{N}(0,I)$. At each reverse diffusion step $i$, a batch of $N_s$ samples of state trajectories is drawn from the Gaussian distribution (line 3, Algorithm \ref{alg:proj-diff})
\begin{equation}
    \tilde{X}^{i} \sim \mathcal{N}\left(\frac{\tilde{x}^i}{\sqrt{\bar{\alpha}_{i-1}}}, \sigma^2_{i,t}I\right), 
    \label{eqn:batch_sampling}
\end{equation}
where the dimension of $\tilde{X}^i$ is ${N_s\times T\times n}$. Then, at the reverse diffusion step $i$, a weighted sample mean (line 5, Algorithm \ref{alg:proj-diff})
\begin{equation}
    \bar{x}^{i} = \frac{\sum_{x \in \tilde{X}^i} x p_J(x)p_g(x)}{\sum_{x \in \tilde{X}^i} p_J(x)p_g(x)}
    \label{eqn:weighted_sample_mean}
\end{equation}
is used to approximate the score function $\nabla_{\tilde{x}^i} \log p_i(\tilde{x}^i)$ (line 6, Algorithm \ref{alg:proj-diff})
\begin{equation}
    \nabla_{\tilde{x}^i} \log p_i(\tilde{x}^i) \approx - \frac{\tilde{x}^i - \sqrt{\bar{\alpha}_i} \, \bar{x}^i}{1 - \bar{\alpha}_i},
    \label{eqn:score_function}
\end{equation}
where $p_i$ results from iteratively adding Gaussian noise to $p_0$. The reverse diffusion process is given by (line 7, Algorithm \ref{alg:proj-diff})
\begin{equation}
    \tilde{x}^{i-1} = \frac{1}{\sqrt{\alpha_i}}(\tilde{x}^i + (1-\bar{\alpha}_i)\nabla_{\tilde{x}^{i}}\log p_i(\tilde{x}^{i})).
    \label{eqn:reverse_diff_ours}
\end{equation}
To address the terminal set constraint $x_T \in \mathcal{X}_T$, we fix the end state by enforcing the boundary condition at $\tilde{x}^i_T:=x_T$ while setting $\sigma_{i,T}=0$. Intuitively, at each reverse diffusion step $i$, our direct sampling approach generates a sequence of states $\tilde{x}^i_{t=1:T}$. 
In \cite{pan2024modelbaseddiffusion}, the authors prove that when $\lambda$ is small enough, the distribution of $J(X)$ with $X\sim p_0(\cdot)$ approaches the optimal cost $J^*$ under mild conditions. 

We now show that the score function~\eqref{eqn:score_function} can indeed be expressed as a linear combination of the predicted state sequence $\tilde{x}^i$ and a weighted sample mean $\bar{x}^i$ in the following proposition.
\begin{proposition}
The score function~\eqref{eqn:score_function} can be approximated as $\nabla_{\tilde{x}^i}\log p_i(\tilde{x}^i) \approx -\frac{-\tilde{x}^i-\sqrt{\bar{\alpha}_i}\bar{x}^i}{1-\bar{\alpha}_i}$, where $\bar{x}^i$ is a weighted sample mean~\eqref{eqn:weighted_sample_mean}.
\label{pro:score}
\end{proposition}
\begin{proof}
The proof is partially adapted from \cite{pan2024modelbaseddiffusion}. We first define the following ``forward process", which iteratively adds Gaussian noise to a target distribution $p_0(\cdot) \propto p_J(\cdot)p_g(\cdot)$:
\begin{equation}
    p_{i|0}(\cdot|\tilde{x}^0) \sim \mathcal{N}\left(\sqrt{\bar{\alpha}_i}\tilde{x}^0, (1-\bar{\alpha}_i)I)\right), \bar{\alpha}_i=\prod_{k=1}^i\alpha_k.
    \label{eqn:forward_gaussian}
\end{equation}
We may also define the forward process at $i$ conditioned on $i-1$, given as:
\begin{equation}
    p_{i|i-1}(\cdot|\tilde{x}^{i-1}) \sim \mathcal{N}\left(\sqrt{\alpha_i}\tilde{x}^{i-1}, (1-\alpha_i)I)\right).
    \label{eqn:forward}
\end{equation}
Our reverse diffusion process~\eqref{eqn:reverse_diff_ours} $p_{i-1|i}(\tilde{x}^{i-1}|\tilde{x}^i)$ stems from the reverse of the forward process $p_{i|i-1}(\tilde{x}^i|\tilde{x}^{i-1})$. From Bayes' rule, we have:
\begin{align}
\nabla_{x^{i}} \log p_i(x^{i}) 
= \frac{\nabla_{\tilde{x}^{i}} \int p_{i|0}\!\left(\tilde{x}^{i} \mid \tilde{x}^{0}\right) 
      p_0(\tilde{x}^{(0)}) \, d\tilde{x}^{(0)}}
      {\int p_{i|0}\!\left(\tilde{x}^{i} \mid \tilde{x}^{0}\right) p_0(\tilde{x}^{0}) \, d\tilde{x}^{0}} 
\\
= \frac{\int \dfrac{-\tilde{x}^{i} + \sqrt{\bar\alpha_i}\,\tilde{x}^{0}}{1-\bar\alpha_i}\,
      p_{i|0}\!\left(\tilde{x}^{i} \mid \tilde{x}^{0}\right) p_0(\tilde{x}^{0}) \, d\tilde{x}^{0}}
      {\int p_{i|0}\!\left(\tilde{x}^{i} \mid \tilde{x}^{0}\right) p_0(\tilde{x}^{0}) \, d\tilde{x}^{0}}
\\
= -\frac{\tilde{x}^{i}}{1-\bar\alpha_i} 
   + \frac{\sqrt{\bar\alpha_i}}{1-\bar\alpha_i}\,
     \frac{\int \tilde{x}^{0} p_{i|0}\!\left(\tilde{x}^{i} \mid \tilde{x}^{0}\right) p_0(\tilde{x}^{0}) \, d\tilde{x}^{0}}
          {\int p_{i|0}\!\left(\tilde{x}^{i} \mid \tilde{x}^{0}\right) p_0(\tilde{x}^{0}) \, d\tilde{x}^{0}},
          \label{eqn:score_estimation_Derivation}
\end{align}
where the forward Gaussian~\eqref{eqn:forward_gaussian} is used to evaluate the gradient. Then, by completing the squares we can show that:
\begin{align}
p_{i|0}(\tilde{x}^i \mid \tilde{x}^0) 
\;\propto\; 
\exp\!\left(
  -\tfrac{1}{2}
  \frac{\bigl(\tilde{x}^{0} - \tfrac{\tilde{x}^i}{\sqrt{\bar\alpha_i}}\bigr)^{\!\top}
        \bigl(\tilde{x}^{0} - \tfrac{\tilde{x}^i}{\sqrt{\bar\alpha_i}}\bigr)}
       {\tfrac{1-\bar\alpha_i}{\bar\alpha_i}}
\right) \label{eqn:complete_square_deriv}\\
\;\propto\; 
\mathcal{N}\!\left(\tfrac{\tilde{x}^i}{\sqrt{\bar\alpha_i}},\;
                 (\tfrac{1}{\bar{\alpha}_i} - 1)I\right).
                 \label{eqn:batch_sample_proof}
\end{align}
Note that~\eqref{eqn:batch_sample_proof} is equivalent to~\eqref{eqn:batch_sampling}, which is our batch sampling step. We substitute~\eqref{eqn:batch_sample_proof} into~\eqref{eqn:score_estimation_Derivation}
and replace the continuous integral term in~\eqref{eqn:score_estimation_Derivation} with Monte Carlo estimation. We treat each $\tilde{x}^0$ in~\eqref{eqn:score_estimation_Derivation} as a sample drawn from~\eqref{eqn:batch_sample_proof} since the integration variable is $\tilde{x}^{0}$. Thus, we obtain:
\begin{equation}
   \nabla_{x^{i}} \log p_i(x^{i}) = -\frac{\tilde{x}^{i}}{1-\bar\alpha_i} 
   + \frac{\sqrt{\bar\alpha_i}}{1-\bar\alpha_i}\,
     \frac{\sum_{x  \in \tilde{X}_i} xp_J(x)p_g(x)}
          {\sum_{x\in \tilde{X}_i} p_J(x)p_g(x)},
\end{equation}
By defining
\begin{equation}
    \bar{x}_i = 
     \frac{\sum_{x \in \tilde{X}_i} xp_J(x)p_g(x)}
          {\sum_{x\in \tilde{X}_i} p_J(x)p_g(x)},
\end{equation}
we are able to establish the result that 
\begin{equation}
    \nabla_{\tilde{x}^i}\log p_i(\tilde{x}^i) \approx -\frac{\tilde{x}^i-\sqrt{\bar{\alpha}_i}\bar{x}^i}{1-\bar{\alpha}_i}.
\end{equation}
\end{proof}
It is noted that dynamic feasibility $p_d(\cdot)$ is not explicitly considered in~\eqref{eqn:weighted_sample_mean} since $p_d(\cdot)$ is enforced by directly applying projection to $\tilde{X}^i$ in~\eqref{eqn:batch_sampling}. We discuss our novel approach for enforcing dynamic feasibility via projection in the following section.

\begin{algorithm}[h]
\caption{Projection Augmented Model-Based Diffusion for Direct Trajectory Optimization}
\label{alg:proj-diff}
\SetKwInOut{Input}{Input}

\Input{$\tilde{x}^{N}\in \mathbb{R}^{T\times n}\sim\mathcal{N}(0,I)$ }

Initialize noise schedule $\sigma_{i,t}$\

\For{$i\leftarrow N$ \KwTo $1$}{
  \textbf{Sample a batch} $$\tilde{X}^{(i)}\sim\mathcal{N}\Bigl(\tfrac{\tilde{x}^{i}}{\sqrt{\bar\alpha_{i-1}}},\,\sigma_{i,t}^2I\Bigr)\;$$
  
  \textbf{Batch projection}: $$\tilde{X}^{(i)}\leftarrow\texttt{project}\bigl(\tilde{X}^{(i)}\bigr)\;$$
  
  \textbf{Compute weighted mean}:
  \[
  \begin{aligned}
    \bar x^{(i)} &= \frac{\sum_{x\in \tilde{X}^{(i)}}x\,p_J(x)p_g(x)}%
                             {\sum_{x\in \tilde{X}^{(i)}}p_J(x)p_g(x)}
  \end{aligned}
  \]
  
  \textbf{Estimate score}:
  \[
    \nabla_{\tilde{x}^{i}}\log p_i(\tilde{x}^{i})
      \approx -\frac{\tilde{x}^{i}-\sqrt{\bar\alpha_i}\,\bar x^{i}}{1-\bar\alpha_i}\;
  \]
  
  \textbf{Update sample}:
  \[
    \tilde{x}^{i-1} 
      = \frac{1}{\sqrt{\alpha_{i}}}\Bigl(\tilde{x}^{i} + (1-\bar{\alpha}_{i})\,\nabla_{\tilde{x}^{i}}\log p_i(\tilde{x}^{i})\Bigr)
  \]
  
  \textbf{Project single sample}:
  \[
    \tilde{x}^{i-1} \leftarrow \texttt{project}\bigl(\tilde{x}^{i-1}\bigr)\;
  \]
}  

\end{algorithm}

\subsection{Dynamic feasibility via gradient-free projection}
\label{sec:methodology_2}
While we have introduced our direct trajectory sampling approach in Section \ref{sec:methodology_1} that gradually generates samples from the target distribution $p_0(\cdot)\sim p_J(\cdot)p_g(\cdot)$, we have not yet accounted for dynamic feasibility $p_d(\cdot)$. Therefore, we augment our approach in Section \ref{sec:methodology_1} by introducing a gradient-free projection to enforce dynamic feasibility recursively to complete our algorithm. Note that the predicted state sequence at reverse diffusion step $i$, i.e., $\tilde{x}^i_{t=1:T}$, is not yet dynamically feasible. Due to the recursive nature of dynamic feasibility, it is necessary to project the predicted state $\tilde{x}^i_{t+1}$ onto the \textit{reachable set} of $\tilde{x}^i_t$. We denote such a reachable set as $\mathcal{H}(\tilde{x}^i_t)$.  We further denote $\mathcal{U}(\cdot)$ as a uniformly random distribution and denote the time-invariant admissible action set as $\mathcal{A}$. For a nonlinear system, we do not have access to the closed-form expression for such a reachable set $\mathcal{H}(\cdot)$.To address this challenge, we adopt a sampling approach: we uniformly draw a batch of $N_p$ action samples, with each sample consisting of all degrees of freedom of the action in $\mathbb{R}^m$. Thus, we obtain $\{U_t\} \in \mathbb{R}^{N_p \times m} \sim \mathcal{U}(\mathcal{A})$, where each action sample $u_t$ in $\{U_t\}$ has $m$ degrees of freedom. The admissible action set $\mathcal{A}:\{u^\zeta_{L} \leq u^\zeta \leq u^\zeta_{U}| \zeta = 1,2,\dots,m\}$ refers to the allowable range of values between the lower bound $u_{L}$ and upper bound $u_{U}$ for each independent degree of freedom $\zeta$. Then, we forward propagate the \textit{predicted current state} $\tilde{x}^i_t$ to obtain a batch of $N_p$ \textit{dynamically feasible samples} by $\{x^i_{t+1}\} = f(\tilde{x}^i_t,u_t)$, for all action samples $u_t \in \{U_t\}$. Next, we replace the \textit{predicted next state} $\tilde{x}^i_{t+1}$ with the feasible sample in $\{x^i_{t+1}\}$ that is closest to $\tilde{x}^i_{t+1}$ in terms of the 2-norm distance. In summary, we project $\tilde{x}^i_{t+1}$ onto $\mathcal{H}(\tilde{x}^i_t)$ by first solving:
\begin{equation}
    u^{*}_t = argmin_{u_t}||f(\tilde{x}^i_t,u_t) - \tilde{x}^i_{t+1}||^2_2, \forall u_t\in \{U_t\}.
    \label{eqn:project_single_State}
\end{equation}
Then, once we find the action $u^{*}_t$ that minimizes the distance $||f(\tilde{x}^i_t,u_t)-\tilde{x}^i_{t+1}||^2_2$, our projected state (dynamically feasible) becomes $x^i_{t+1} = f(\tilde{x}^i_t,u^*_t)$, which completes the projection step. This process then repeats for all $t=0,1,\dots,T-2$. Our complete reverse diffusion process is formulated in Algorithm \ref{alg:proj-diff}. 
\begin{figure*}[!t]
  \centering
  \includegraphics[width=\linewidth]{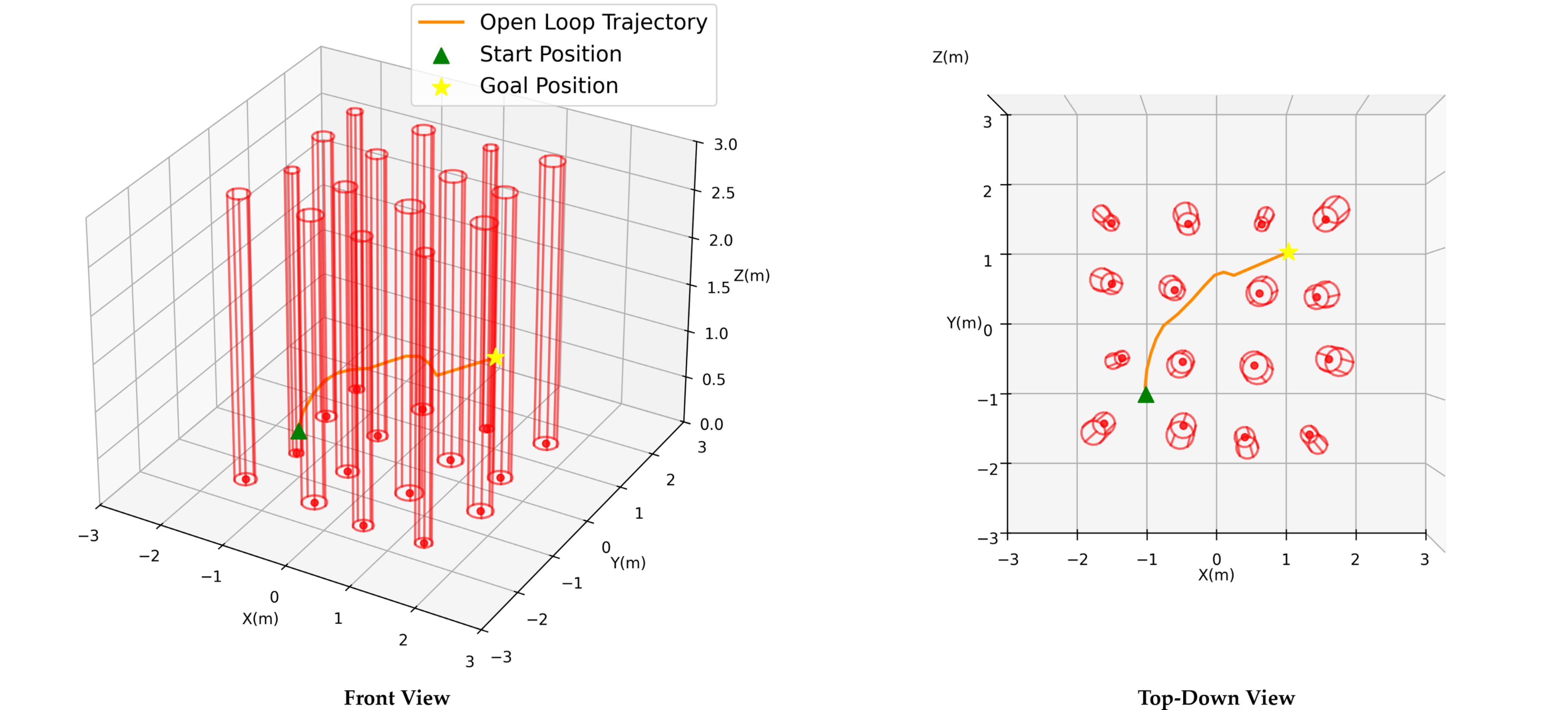}
  \caption{Front view and top-down view of open-loop trajectory generation via Alg. \ref{alg:proj-diff} for a generic waypoint navigation scenario.}
  \label{fig:test_trial}
\end{figure*}

In addition, as the authors have noted in \cite{Bouvier2025DDATDP}, it is only meaningful to start projecting the states when the noise level $\sigma_{i,t}$ is low. This is due to the fact that during the initial exploration phase with high noise, the sampled trajectories are far away from optimality. Empirically, it is also beneficial to randomly choose which time horizon $t$ to project $\tilde{x}^i_t$, for any diffusion steps $i$, instead of projecting at all the time horizons. This partial projection method has the effect of minimizing the overall projection error and speeding up the projection process. We show our projection schedule in Algorithm \ref{alg:proj_schedule}, where we adopt a linearly interpolated projection probability between two noise thresholds $\sigma_{min}$ and $\sigma_{max}$, similar to \cite{Bouvier2025DDATDP}. Since our noise schedule $\sigma_{i,t}\in \mathbb{R}^{T \times N}$ spans both the trajectory prediction horizon and the diffusion horizon, we use the average of the noise schedule computed along the trajectory prediction horizon to determine the projection threshold, i.e., $mean(\sigma_{i})=\frac{\sum^{k=T}_{k=1}\sigma_{i,k}}{T}$.
\begin{algorithm}[h]
\caption{Projection Schedule Conditioned on Noise}
\label{alg:proj_schedule}
\SetKwInOut{Input}{Input}

\Input{$x^i$ or $X^i$}
Choose parameters $\sigma_{\max}$ and $\sigma_{\min}$\;

\If{mean($\sigma_i) > \sigma_{\max}$}{
    no projection\;
}
\ElseIf{$\sigma_{\min} \leq mean(\sigma_i) \leq \sigma_{\max}$}{
    Compute projection probability:
    \[
      p_{\text{proj}} = 
      \mathrm{clip}\!\left(
      \tfrac{\sigma_{\max}-mean(\sigma_i)}{\sigma_{\max}-\sigma_{\min}},0,1\right)
    \]
    
    For each timestep $t$ in $x^i$ or $X^i$:
    \begin{enumerate}
      \item Draw a random number $n_t \sim \mathcal{U}(0,1)$
      \item If $n_t < p_{\text{proj}}$, project $X^i_t$ or $x^i_t$
      \item Otherwise, leave $X^i_t$ or $x^i_t$ unchanged
    \end{enumerate}

}
\Else{
    always project\;
}
\end{algorithm}
Our projection scheme differs from \cite{Bouvier2025DDATDP} in that our approach preserves the gradient-free nature of sampling-based trajectory optimization, whereas \cite{Bouvier2025DDATDP} requires solving a convex optimization problem at each projection step.
\begin{remark}
Our bi-level noise schedule $\sigma_{i,t}$ differs from the one used in \cite{dialMPC}. In \cite{dialMPC}, the noise along the trajectory prediction horizon $t=0,1,\dots,T$ increases as $t$ increases, allowing for more exploration of control inputs in the later time steps; in contrast, our algorithm diffuses states instead of control inputs. We deliberately choose the noise to decrease as $t$ increases. Intuitively, this choice adapts the bi-level noise schedule to our novel projection mechanism (see Section \ref{sec:methodology_2}), which only enforces dynamic feasibility when we obtain high-quality samples in the reverse diffusion process, i.e., when the noise level is low. We empirically discover that this leads to smoother trajectories. 
\end{remark}

%% file: Experiments/Results.tex
We compare our projection-augmented diffusion for direct trajectory optimization with MBD\cite{pan2024modelbaseddiffusion}, DRAX \cite{kurtz2024equalityconstraineddiffusion}, and an NLP solver, Casadi\cite{casadi}. 

\begin{table*}[!t]
\centering
\caption{Performance Comparison of Open-Loop Planning}
\label{tab:performance}
\begin{tabular}{lccccccc}
\toprule
\textbf{Method} &
\textbf{Success(\%)} &
\textbf{Dist.\ to Goal[$m$]} &
\textbf{Clearance[$m$]} &
\textbf{Length[$m$]} &
\textbf{Time[$s$]} &
\textbf{Dynamic Feasibility Error} &
\\
\midrule
MBD($N$=200) & 68.0 & $0.6008 \pm 0.5646$ & $0.1175 \pm 0.1723$ & $4.5907 \pm 2.4694$ & $ 14.35 \pm 1.19$ & $0 \pm 0$  \\
DRAX($N$=200) & 21.0 & $0.1521 \pm 0.0376$ & $-0.0583 \pm 0.0864$ & $4.5299\pm 0.9942$ & $8.30 \pm 0.55$ & $3.3065 \pm 2.3365$\\
DRAX($N$=400)& 24.0 & $0.1093 \pm 0.0316$ & $-0.0580 \pm 0.0932$& $4.3500 \pm 1.0588$ &  $8.63 \pm 0.68$ & $4.4727 \pm 6.2674$\\
NLP & 53.0 & $0.0663 \pm 0.0360$ & $0.1077 \pm 0.2583$ & $4.4426 \pm 1.4822$ & $12.62 \pm 1.76 $ & $0 \pm 0$\\
\textbf{Ours($N$=200)} & 78.0 & $0 \pm 0$ & $0.0689 \pm 0.1050$ & $4.2713 \pm 2.0723$ & $18.36 \pm 6.57 $ & $0 \pm 0$\\

\bottomrule
\end{tabular}
\end{table*}


\subsection{Waypoint Navigation Scenario}
\label{sec:waypoint_navigation}
In our simulation studies, we focus on a quadrotor system with the following dynamics \cite{geometric_se3}:
\begin{align}
  \dot{o} &= v \label{eq:xdot} \\[6pt]
  m\dot{v} &= m g e_3 - F R e_3 \label{eq:vdot} \\[6pt]
  \dot{R} &= R \hat{\Omega} \label{eq:Rdot} \\[6pt]
\Gamma \dot{\Omega} + &\Omega \times \Gamma\Omega = M \label{eq:Omegadot}
\end{align}
where $o=[o_x,o_y,o_z] \in \mathbb{R}^3$ refers to the position vector, $v=[v_x,v_y,v_z]\in \mathbb{R}^3$ refers to the velocity vector, $R$ refers to the rotation matrix, and $\Omega = [\omega_x,\omega_y,\omega_z] \in \mathbb{R}^3$ refers to the angular velocity vector, $\Gamma$ refers to the inertia matrix. Additionally, the control inputs are the thrust along the vertical body axis, denoted by $F$, and the torque vector, denoted by $M  = [M_x, M_y, M_z] \in \mathbb{R}^3$. The hat map $\hat{\cdot}$ maps a vector in $\mathbb{R}^3$ to a 3 by 3 skew-symmetric matrix representation. We assume the quadrotor has a mass of 1 $kg$, and a diagonal inertial matrix with $\Gamma_x$ = 0.01 $kg \cdot m^2$, $\Gamma_y$ = 0.01 $kg \cdot m^2$, and $\Gamma_z$ = 0.02 $kg \cdot m^2$. We showcase the effectiveness of our Algorithm \ref{alg:proj-diff} in a waypoint navigation scenario. We assume a workspace of dimensions $6 m \times 6m \times 3m$. We place 16 static cylindrical obstacles in the workspace with variable radii, randomly sampled from 0.1 meters to 0.2 meters. We assume that each cylindrical obstacle has the same height of 3.0 meters, spanning from the ground to the ceiling. By doing so, we only need to consider the horizontal distance between the quadrotor to the surface of each cylinder in the safety constraints $g(\cdot)$~\eqref{eqn:collision_avoidance}. 

We first validate Algorithm \ref{alg:proj-diff} on a \textbf{single test case}, then we comprehensively compare our approach with baselines over \textbf{100 randomized trials} (see Section \ref{sec:monte_Carlo_comparisons}). In the single test case, we assume the initial position is given by $[-1,-1, 0.5] \in \mathbb{R}^3$ and the goal position is given by $[1, 1, 1]$ $\in \mathbb{R}^3$. All positional units are measured in meters. We use a trajectory prediction horizon of $T$ = 50, a sampling interval of $dt$ = 0.1 $s$, a diffusion horizon of $N$ = 200, a temperature $\lambda$ = 0.1 in $p_J(\cdot)$, and $\kappa = 5$ in $p_g(\cdot)$. We apply the Runge-Kutta 4 (RK4) discretization method to simulate the dynamics. We use $N_s$ = 256 samples for the batch sampling step in Algorithm \ref{alg:proj-diff}. For the projection mechanism, we use $N_p$ = 250 samples, $\sigma_{max}=0.3$, and $\sigma_{min}=0.1$. For our noise schedule~\eqref{eqn:noise_Schedule_ours}, we use $\beta_0$ = $1\times10^{-4}$ and $\beta_N = 1\times10^{-2}$, which determines $\bar{\alpha}$. We also choose $\delta = 0.8$~\eqref{eqn:noise_Schedule_ours}, which controls the noise level over the trajectory prediction horizon. We show the open-loop trajectory in Figure \ref{fig:test_trial}.

\begin{figure*}[!t]
  \centering
  \includegraphics[width=\linewidth]{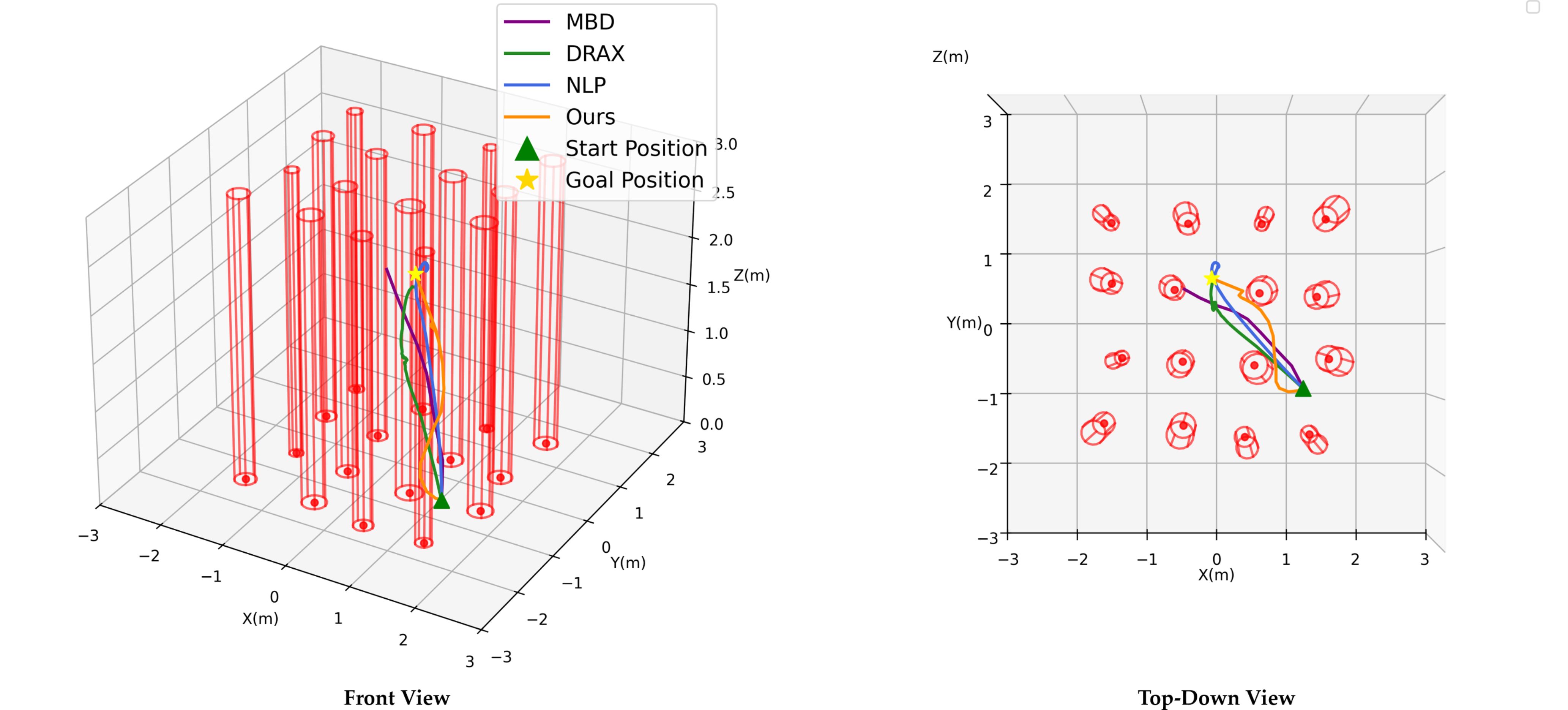}
  \caption{Comparison of open-loop trajectories generated by our method and the baselines. All trajectories shown here are collision-free.}
  \label{fig:comparison_combined}
\end{figure*}

\subsection{Comparison Against Baselines}
\label{sec:monte_Carlo_comparisons}
We comprehensively validate our approach in open-loop trajectory by comparing it to three baselines: MBD \cite{pan2024modelbaseddiffusion}, DRAX \cite{kurtz2024equalityconstraineddiffusion}, and NLP \cite{casadi} in the same workspace defined in Section \ref{sec:waypoint_navigation}. In each simulation trial, the initial and goal position pairs are randomly generated outside the obstacles and are at least 0.1 $m$ away from the surface of each obstacle. To be consistent with MBD \cite{pan2024modelbaseddiffusion} and DRAX \cite{kurtz2024equalityconstraineddiffusion}, our implementation also relies on the JAX package \cite{jax2018github}, which enables GPU accelerated parallel computations in the forward propagation of the dynamics. All simulations are performed on a PC with an NVIDIA RTX 4080 GPU with 16GB of RAM.

We evaluate the performance of open-loop trajectory planning. We present the statistics collected from 100 randomized trials for open-loop trajectory optimization in Table \ref{tab:performance}. We define the success rate as the percentage of trials where no collision occurs. To evaluate the dynamic feasibility of our Algorithm \ref{alg:proj-diff} after the diffusion process is complete at $i=1$~\eqref{eqn:reverse_diff_ours}, we apply the control input $u_t$ that is chosen during projection~\eqref{eqn:project_single_State}. To evaluate the dynamic feasibility of DRAX \cite{kurtz2024equalityconstraineddiffusion}, we apply the control inputs returned by their algorithm. We drop the $\tilde{\cdot}$ on $x^0_t$ here since the reverse diffusion process is complete, and our projection mechanism is applied to all states along the trajectory prediction horizon $t$ towards the end of the reverse diffusion process (see Algorithm \ref{alg:proj_schedule}). We define the trajectory-wise dynamic feasibility error as:
\begin{equation}
    err_{f} = \frac{\sum^{T-1}_{t=0}||f(x_t,u_t) - x_{t+1}||^2_2}{T}.
    \label{eqn:dynamics_Err}
\end{equation}

We define the success of open-loop planning as a trajectory that leads to no collision with the boundary of the workspace or the obstacles. From Table \ref{tab:performance}, we observe that in open-loop trajectory planning, our method outperforms all baselines in terms of the success rate, distance to goal, and trajectory length. However, the robustness of our method comes at the cost of extra computation time compared to the other baselines. While DRAX \cite{kurtz2024equalityconstraineddiffusion} also applies direct trajectory optimization via a diffusion-style process, it results in a high dynamic feasibility error, whereas our approach leads to \textbf{zero dynamic feasibility error}. In addition, our method leads to approximately \textbf{4x higher success rate} compared to DRAX \cite{kurtz2024equalityconstraineddiffusion}. We note that the average clearance from obstacles for DRAX \cite{kurtz2024equalityconstraineddiffusion} is negative, indicating that it is highly prone to collisions\footnote{Clearance distance is positive if a trajectory lies outside the surface of the obstacles; otherwise, it is negative.}. While MBD \cite{pan2024modelbaseddiffusion} yields a success rate of 68\% in open-loop planning, it leads to sub-optimal trajectories as reflected by the trajectory length and the distance to goal error, i.e., the distance between the position at the last time step $T$ of the open-loop trajectory and the designated goal position. We observe that NLP \cite{casadi} leads to a 25\% lower success rate than our Algorithm \ref{alg:proj-diff}, indicating its lack of robustness.

Regarding the computation time, when using the same number of diffusion steps ($N$=200), we observe that DRAX \cite{kurtz2024equalityconstraineddiffusion} outperforms our method at the cost of a \textbf{3x lower} success rate and non-zero dynamic feasibility error $err_f=3.3 \pm 2.3$. In addition, when we double the number of diffusion steps for DRAX ($N$=400), the success rate barely improves. In addition, this leads to an even higher dynamic feasibility error overall. Intuitively, DRAX \cite{kurtz2024equalityconstraineddiffusion} is unable to sample from the true target distribution of trajectories that account for both optimality and dynamic feasibility. We also note that NLP \cite{casadi}, MBD \cite{pan2024modelbaseddiffusion}, and DRAX \cite{kurtz2024equalityconstraineddiffusion} are all faster than our approach. While DRAX \cite{kurtz2024equalityconstraineddiffusion} is twice as fast as our approach, we find that MBD \cite{pan2024modelbaseddiffusion} is comparable to ours. The relatively slow computation speed is attributed to our projection mechanism, which cannot be parallelized due to its chronological nature. Further study is required to investigate how we can speed up our projection-augmented diffusion framework. Finally, we show a visual comparison from one of the simulation trials between the trajectories generated by our method and the baselines in Figure \ref{fig:comparison_combined}. We observe that MBD \cite{pan2024modelbaseddiffusion} misses the goal position and almost collides with an obstacle. Trajectories generated by DRAX \cite{kurtz2024equalityconstraineddiffusion}, NLP \cite{casadi}, and our approach all reach the goal position. For NLP, the trajectory initially overshoots the goal position before reaching it, indicating sub-optimality. Our approach leads to a trajectory that has more curvature than DRAX \cite{kurtz2024equalityconstraineddiffusion} since there is no dynamic feasibility violation.

%% file: references.bib
@inproceedings{
zhang2025constraineddiffusers,
title={Constrained Diffusers for Safe Planning and Control},
author={Jichen Zhang and Liqun Zhao and Antonis Papachristodoulou and Jack Umenberger},
booktitle={The Thirty-ninth Annual Conference on Neural Information Processing Systems},
year={2025},
url={https://openreview.net/forum?id=tahkGZjjWA}
}

@inproceedings{
pan2024modelbaseddiffusion,
title={Model-based Diffusion for Trajectory Optimization},
author={Chaoyi Pan and Zeji Yi and Guanya Shi and Guannan Qu},
booktitle={The Thirty-eighth Annual Conference on Neural Information Processing Systems},
year={2024},
url={https://openreview.net/forum?id=BJndYScO6o}
}

@misc{kurtz2024equalityconstraineddiffusion,
  title        = {Equality Constrained Diffusion for Direct Trajectory Optimization},
  author       = {Kurtz, Vince and Burdick, Joel W.},
  year         = {2024},
  eprint       = {2410.01939},
  archivePrefix= {arXiv},
  primaryClass = {cs.RO},
  url          = {https://arxiv.org/abs/2410.01939}
}

@inproceedings{Ho2020DenoisingDP,
  title={Denoising diffusion probabilistic models},
  author={Ho, Jonathan and Jain, Ajay and Abbeel, Pieter},
  booktitle={Advances in Neural Information Processing Systems},
  volume={33},
  pages={6840--6851},
  year={2020}
}

@Article{casadi,
  author = {Joel A E Andersson and Joris Gillis and Greg Horn
            and James B Rawlings and Moritz Diehl},
  title = {{CasADi} -- {A} software framework for nonlinear optimization
           and optimal control},
  journal = {Mathematical Programming Computation},
  volume = {11},
  number = {1},
  pages = {1--36},
  year = {2019},
  publisher = {Springer},
  doi = {10.1007/s12532-018-0139-4}
}

@conference{Bouvier2025DDATDP,
author={ Jean-Baptiste Bouvier and Kanghyun Ryu and Kartik Nagpal and Qiayuan Liao and Koushil Sreenath and Negar Mehr },
title={ {DDAT}: Diffusion Policies Enforcing Dynamically Admissible Robot Trajectories },
booktitle={ Robotics: Science and Systems (RSS) },
pages={ },
month={ June },
year={ 2025 },
address={ },
}

@article{MPPI,
  title={Model predictive path integral control using covariance variable importance sampling},
  author={Williams, Grady and Aldrich, Andrew and Theodorou, Evangelos},
  journal={arXiv preprint arXiv:1509.01149},
  year={2015}
}

@article{dialMPC,
  title={Full-Order Sampling-Based MPC for Torque-Level Locomotion Control via Diffusion-Style Annealing},
  author={Haoru Xue and Chaoyi Pan and Zeji Yi and Guannan Qu and Guanya Shi},
  journal={2025 IEEE International Conference on Robotics and Automation (ICRA)},
  year={2024},
  pages={4974-4981},
  url={https://api.semanticscholar.org/CorpusID:272832507}
}

@article{diffusion_policy,
  title={Diffusion policy: Visuomotor policy learning via action diffusion},
  author={Chi, Cheng and Xu, Zhenjia and Feng, Siyuan and Cousineau, Eric and Du, Yilun and Burchfiel, Benjamin and Tedrake, Russ and Song, Shuran},
  journal={The International Journal of Robotics Research},
  pages={02783649241273668},
  year={2023},
  publisher={SAGE Publications Sage UK: London, England}
}

@inproceedings{safe_diffuser,
  title={Safediffuser: Safe planning with diffusion probabilistic models},
  author={Xiao, Wei and Wang, Tsun-Hsuan and Gan, Chuang and Hasani, Ramin and Lechner, Mathias and Rus, Daniela},
  booktitle={The Thirteenth International Conference on Learning Representations},
  year={2023}
}

@INPROCEEDINGS{altro,
  author={Howell, Taylor A. and Jackson, Brian E. and Manchester, Zachary},
  booktitle={2019 IEEE/RSJ International Conference on Intelligent Robots and Systems (IROS)}, 
  title={ALTRO: A Fast Solver for Constrained Trajectory Optimization}, 
  year={2019},
  volume={},
  number={},
  pages={7674-7679},
  doi={10.1109/IROS40897.2019.8967788}}

@article{CEM,
author = {Kobilarov, Marin},
year = {2012},
month = {05},
pages = {855-871},
title = {Cross-entropy motion planning},
volume = {31},
journal = {International Journal of Robotic Research - IJRR},
doi = {10.1177/0278364912444543}
}

@INPROCEEDINGS{geometric_se3,
  author={Lee, Taeyoung and Leok, Melvin and McClamroch, N. Harris},
  booktitle={49th IEEE Conference on Decision and Control (CDC)}, 
  title={Geometric tracking control of a quadrotor UAV on SE(3)}, 
  year={2010},
  volume={},
  number={},
  pages={5420-5425},
  doi={10.1109/CDC.2010.5717652}}

@INPROCEEDINGS{sequential_convex_fleet,
  author={Augugliaro, Federico and Schoellig, Angela P. and D'Andrea, Raffaello},
  booktitle={2012 IEEE/RSJ International Conference on Intelligent Robots and Systems}, 
  title={Generation of collision-free trajectories for a quadrocopter fleet: A sequential convex programming approach}, 
  year={2012},
  volume={},
  number={},
  pages={1917-1922},
  doi={10.1109/IROS.2012.6385823}}

@ARTICLE{quadrotor_generation_gao_Fei,
  author={Zhou, Boyu and Gao, Fei and Wang, Luqi and Liu, Chuhao and Shen, Shaojie},
  journal={IEEE Robotics and Automation Letters}, 
  title={Robust and Efficient Quadrotor Trajectory Generation for Fast Autonomous Flight}, 
  year={2019},
  volume={4},
  number={4},
  pages={3529-3536},
  doi={10.1109/LRA.2019.2927938}}

@article{Byrd1999AnIP,
  title={An Interior Point Algorithm for Large-Scale Nonlinear Programming},
  author={Richard H. Byrd and Mary E. Hribar and Jorge Nocedal},
  journal={SIAM J. Optim.},
  year={1999},
  volume={9},
  pages={877-900},
  url={https://api.semanticscholar.org/CorpusID:16293345}
}

@article{LaValle1998RapidlyexploringRT,
  title={Rapidly-exploring random trees : a new tool for path planning},
  author={Steven M. LaValle},
  journal={The annual research report},
  year={1998},
  url={https://api.semanticscholar.org/CorpusID:14744621}
}

@inproceedings{
DDIM,
title={Denoising Diffusion Implicit Models},
author={Jiaming Song and Chenlin Meng and Stefano Ermon},
booktitle={International Conference on Learning Representations},
year={2021},
url={https://openreview.net/forum?id=St1giarCHLP}
}

@inproceedings{langevin_sampling,
  title={Bayesian Learning via Stochastic Gradient Langevin Dynamics},
  author={Max Welling and Yee Whye Teh},
  booktitle={International Conference on Machine Learning},
  year={2011},
  url={https://api.semanticscholar.org/CorpusID:2178983}
}

@software{jax2018github,
  author = {James Bradbury and Roy Frostig and Peter Hawkins and Matthew James Johnson and Chris Leary and Dougal Maclaurin and George Necula and Adam Paszke and Jake Vander{P}las and Skye Wanderman-{M}ilne and Qiao Zhang},
  title = {{JAX}: composable transformations of {P}ython+{N}um{P}y programs},
  url = {http://github.com/jax-ml/jax},
  version = {0.3.13},
  year = {2018},
}

@article{kelly_trajectory_opt_tutorial,
  title={An Introduction to Trajectory Optimization: How to Do Your Own Direct Collocation},
  author={Matthew Kelly},
  journal={SIAM Rev.},
  year={2017},
  volume={59},
  pages={849-904},
  url={https://api.semanticscholar.org/CorpusID:33541404}
}
